\def\year{2016}\relax
\documentclass[letterpaper]{article}
\usepackage{aaai16}
\usepackage{times}
\usepackage{helvet}
\usepackage{courier}
\usepackage{algorithm}
\usepackage{algorithmic}
\usepackage{enumitem}
\usepackage{subfigure}
\usepackage{float}
\usepackage{color}
\usepackage{pdfpages}
\usepackage{epstopdf}
\usepackage{url}
\newfloat{algorithm}{t}{lop}
\usepackage[english]{babel}
\usepackage{amsmath,amssymb,graphicx,bbm,bm}

\newcounter{eqn}

\makeatletter
\newcommand{\putindeepbox}[2][0.7\baselineskip]{{%
    \setbox0=\hbox{#2}%
    \setbox0=\vbox{\noindent\hsize=\wd0\unhbox0}
    \@tempdima=\dp0
    \advance\@tempdima by \ht0
    \advance\@tempdima by -#1\relax
    \dp0=\@tempdima
    \ht0=#1\relax
    \box0
}}
\makeatother

\begin{document}

\title{Learning a Hybrid Architecture for Sequence Regression and Annotation}
\author{}
\author{Yizhe Zhang \and Ricardo Henao \and Lawrence Carin \\ Department of Electrical and Computer Engineering, Duke University, Durham, NC 27708, USA
 \\ \AND Jianling Zhong \and Alexander J. Hartemink\\ Program in Computational Biology and Bioinformatics, Duke University, Durham, NC 27708, USA}
\maketitle

\begin{abstract}
When learning a hidden Markov model (HMM), sequential observations can often be complemented by real-valued summary response variables generated from the path of hidden states. Such settings arise in numerous domains, including many applications in biology, like motif discovery and genome annotation. In this paper, we present a flexible framework for jointly modeling both latent sequence features and the functional mapping that relates the summary response variables to the hidden state sequence. The algorithm is compatible with a rich set of mapping functions. Results show that the availability of additional continuous response variables can simultaneously improve the annotation of the sequential observations and yield good prediction performance in both synthetic data and real-world datasets.
\end{abstract}

\section{Introduction}
The hidden Markov model (HMM) is a probabilistic formalism for reasoning about sequences of random variables, which naturally arise in temporal (e.g., in natural language) and spatial settings (e.g., in images and biological sequences). HMMs are especially valued for their efficient inference, broad applicability, and rich extensibility. In the traditional context, an HMM enables one to reason about a sequence of unobserved state variables that evolve under a Markov assumption, based only on a sequence of observations corresponding to the hidden state sequence \cite{Bishop:2006ui}. Here, we extend this model to incorporate additional side information. Specifically, we assume that in addition to the sequential observations, we have access to a corresponding vector of real-valued ``response variables" that somehow provide summary information about the hidden state sequence (if this vector is empty, we have a traditional HMM). We are interested in the following dual task: simultaneously learning the \emph{hidden state sequence} from the sequential observations and summary response variables (sequence annotation), as well as the \emph{mapping function} that produces real-valued response variables from the hidden state sequence (sequence regression).

Although applications exist in many domains, this extension to a traditional HMM was originally motivated by a large collection of biological applications. For example, promoter DNA sequences can be complemented by promoter activity to learn a model that annotates the DNA sequence, and simultaneously predicts the activity level of putative promoters. In this paper, we focus on another question: understanding protein-DNA binding preferences, which is also important in unraveling gene regulation. We are particularly interested in learning how transcription factors (TFs) exhibit sequence binding preferences (motifs), in our case modeled as position weight matrices (PWMs) \cite{Berg:1987wu}. In situations where bound genomic DNA sequences are available, HMMs have been applied to jointly learn binding locations and motifs in an unsupervised manner \cite{Bailey:1994tg,Lawrence:1993jg}. However, newer protein binding microarray (PBM) technology allows us to learn motifs from \emph{in vitro} sequence data \cite{Mukherjee:2004iv}. In a typical PBM, about 40,000 synthetic DNA sequences (probes) are placed on a microarray, and the binding affinity of a TF for each probe is reported in the form of a real-valued fluorescence intensity. Thus, motif inference on PBM data requires jointly modeling DNA sequences and real-valued measurements, which poses new challenges for model design.

Many algorithms have been developed for analyzing PBM data. Approaches based on K-mers (K-grams of DNA sequence) feed occurrences of each K-mer as features into more sophisticated models \cite{Weirauch:2013ju,Berger:2006fv}. RankMotif++ \cite{Chen:2007ed} maximizes the likelihood for binarized preference observations. Other approaches attempt to jointly learn a probabilistic model. For example, BEELM-PBM \cite{Zhao:2011isa} estimates parameters based on a biophysical energy model. One advantage of probabilistic models over K-mer models is that binding preferences can be directly described via PWMs. DeepBind \cite{Alipanahi:2015fba} uses a deep convolutional neural network to improve accuracy. A systematic survey of several deterministic and probabilistic models was provided by Weirauch et al. \cite{Weirauch:2013ju}. In a PBM scenario, the observed responses reflect the average intensity level over time and probe instances, i.e., a cluster of probes with same DNA sequence. Probabilistic frameworks such as HMMs can represent the probability distribution over all possible binding configurations, and thus may have advantages over deterministic methods. 

In this paper, we develop inference methods, which we call RegHMM, that efficiently estimate both the hidden state sequence and all the various parameters of the regression-coupled HMM model. The highlights are: 
%
\begin{itemize}
  \item A probabilistic generative model that jointly considers sequential observations and multiple real-valued responses.
  \item Customized link functions, summary functions, and position-wise transition probabilities providing rich and flexible functional mapping.
  \item Fast and efficient inference with approximate expectation-maximization via Viterbi path integration.
  \item Our methodology enables prediction at least as accurate as any other approach to date, while learning a much more interpretable model.
\end{itemize}

\section{Related Work}
Many methods have considered the task of mapping sequential observations to categorical or real responses. Discriminative HMMs use Bayesian model comparison by cross-entropy for sequence classification. Previous algorithms \cite{Collins:2002hh,BenYishai:2004du,Sak:2014vt} investigated discriminative training of HMM or recurrent neural networks (RNN) using maximum mutual information (MMI). \cite{Srivastava:2007cr} uses a profile HMM to classify biological sequences. However, many of these methods cannot straightforwardly extend to regression problems. Kernel methods \cite{Leslie:2002tx} have also been devised to map sequential data to categorical or real responses. However, these kernel methods usually employ predefined kernels and the learned latent features encoding the original sequences are less interpretable than those coming from HMMs. Our approach, in contrast, can be applied for both classification and regression tasks and directly learn the hidden features explaining the sequences.

Previous research has also examined the use of continuous side information to complement sequential observation when learning latent features. Input-output HMMs \cite{Bengio:1996tl} and TRBMs \cite{Zeiler:2011vm} consider the task of supervised learning where the observed variables influence both latent and output variables. However, these approaches assume that input observations for each time point are known. GCHMMs \cite{Fan:2015fy} consider an HMM where transition probabilities are associated with side information. CRBMs \cite{Taylor:2009tt} incorporate side information into the factorization of weights in conditional RBMs. The RUPA method \cite{Noto:2008uq} learns an HMM with additional responses via augmented visit aggregation. However, point estimate approximation over expected visits ignores the true distribution of visits. As a result, convergence problems may arise, especially when the distribution of visits has multiple modes. In contrast, our method uses Viterbi path integration to directly approximate the true posterior of visits. In summary, we consider the responses as both extra information for learning a latent representation of sequential observation, and a variable to predict. Considering the dual task simultaneously enforces interpretability, while maintaining predictive power.

\section{Learning a coupled regression HMM model}
\subsection{Learning task}
\begin{figure}[t!]
	\centering
	\resizebox{200px}{!}{\includegraphics{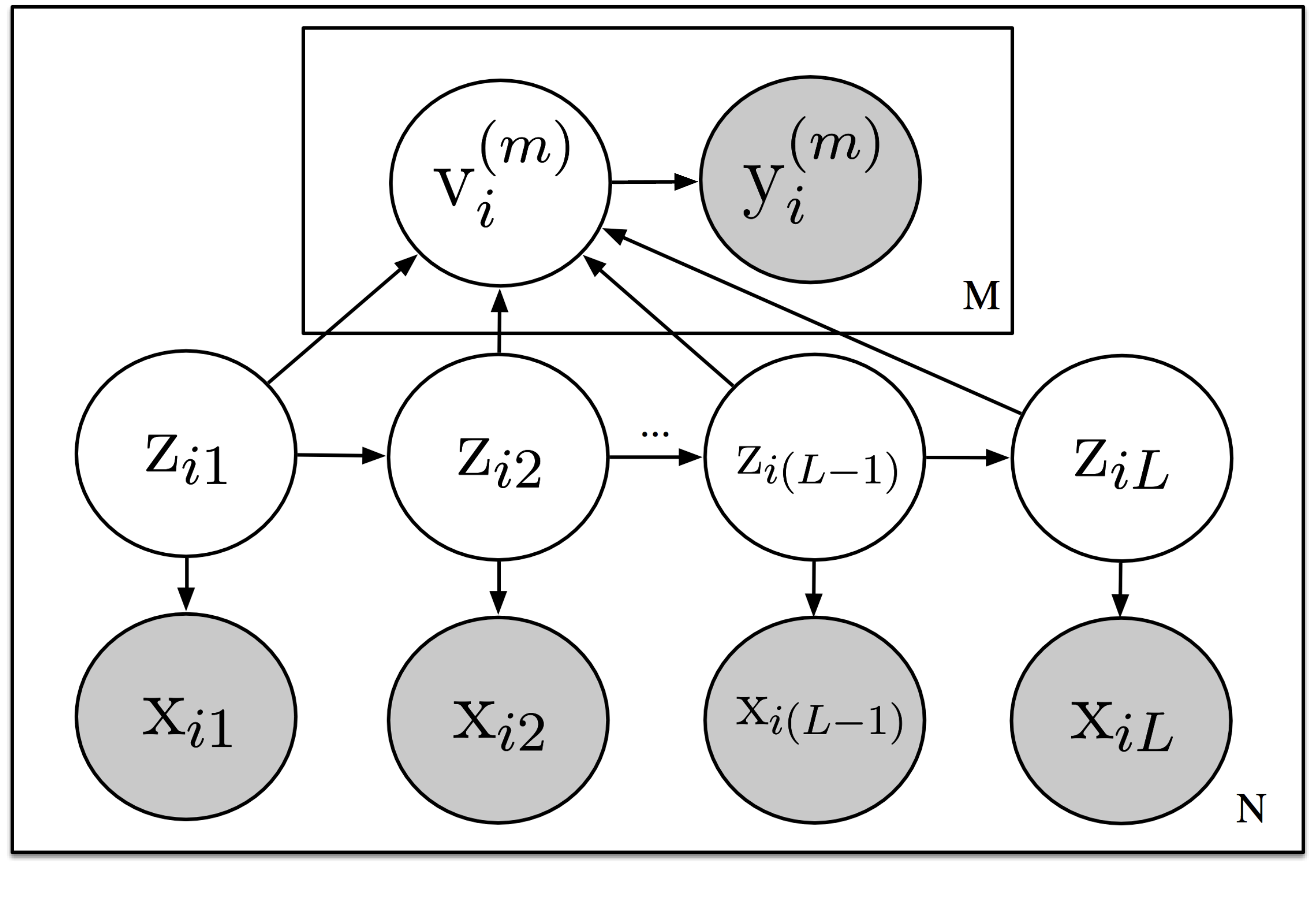}}
	\caption{Graphical model representation of RegHMM.}\label{fig1}
\end{figure}

We propose the RegHMM (regression-coupled HMM), which can be understood as a combination of HMM and regression. As shown in Figure~\ref{fig1}, suppose we have $N$ sequence observations of length $L$, we denote $N\times L$ matrices: $\bm{X}=(\bm{X}_{1},\ldots,\bm{X}_{N})$ to be the sequential observations and $\bm{Z}$ to be the hidden states. An HMM is defined by parameters $\bm{\Theta} = \{\bm{E},\bm{A},\bm{\pi\}}$, where $\bm{E}$ is the emission matrix, $\bm{A}$ is the transition matrix and $\bm{\pi}$ is the vector of initial probabilities. Further, the hidden variables, $\bm{Z}$, of the HMM are employed as covariates for $M$ regression tasks based on the assumption that the real-valued responses, $\bm{Y}=(\bm{Y}^{(1)},\ldots,\bm{Y}^{(M)})$, are generated from summary statistics, $\bm{V}=(\bm{V}^{(1)},\ldots,\bm{V}^{(M)})$, that encode the information of hidden state paths, where $\bm{Y}^{(m)} = ( y^{(m)}_{1} , \ldots ,y^{(m)}_{N} )$ and $\bm{V}^{(m)} = ( v^{(m)}_{1} , \ldots ,v^{(m)}_{N} )$.

As a generative process, the $m$-th response for $i$-th sequential observation, $y^{(m)}_{i}$, is drawn from hidden states encoded by hidden variables $\bm{Z}\in\{1\ldots S\}^{N\times L}$, where $S$ is the total number of discrete states. This generative process of responses can be formally represented as,
\begin{eqnarray*}
  y_{i}^{( m )} & \sim & \mathcal{N} ( \mu_{i}^{( m )} , \sigma^{( m )} ) \,, \\
  \mu_{i}^{( m )} & = & f^{(m)} ( v^{( m )}_{i} ; \bm{\Gamma}^{(m)} ) \,, \\
  v_{i}^{( m )} & = & g^{(m)} ( \bm{Z}_{i} ) \,,
\end{eqnarray*}
where $v_{i}^{( m )}$ first summarize all hidden states, $\bm{Z}_{i} = ( z_{i1},\ldots,z_{iL} )$, via certain summary function, such as counting the visits of certain states. These hidden states summaries, $v_{i}^{(m)}$, are then transformed through a link function, yielding response means $\mu_{i}^{( m )}$. Finally, $y^{( m )}_{i}$ is drawn from a Gaussian distribution with mean $\mu_{i}^{( m )}$. The standard error  $\sigma^{(m)}$, is shared across samples and learned from data. We let $f^{(m)}(\cdot)$ and $g^{(m)}(\cdot)$ to denote the link function and the summary function, respectively. The link function can take either linear form, $f^{(m)} ( v ) = \alpha + \beta v$ or a generalized hyperbolic tangent form (tanh), $f^{(m)} ( v; \alpha , \beta ,s,t )  =  \alpha + \frac{\beta}{1+ \exp [ -s ( v-t ) ]}$, where $\alpha$ and $\beta$ characterize the intercept and slope of $f^{(m)}(\cdot)$, while $s$ and $t$ characterize its shape and center position. Further, Let $\bm{\Gamma}^{(m)}$ denote all parameters defining the link function, i.e., $\{\alpha, \beta, s, t\}$ for the $m$-th task.

The learning aim is to jointly estimate regression parameters, $\bm{\Gamma}^{(m)}$, describing the link functions, as well as the HMM parameters, $\bm{\Theta}$.

\subsection{Analyzing PBM data}
In this section, we consider an application of RegHMM to PBM data analysis. In particular, we desire to infer the emission matrix, representing the binding preference, from probe sequences $\bm{X} \in \{ A,C,G,T \}^{N \times L}$ and real responses $\bm{Y} \in \mathbb{R}^{N}$ where each sequential observation $\bm{X_i}$, correspond to a single response. Here and throughout the rest of this paper, we only discuss the case of a single response task ($M=1$). Thus, for simplicity the subscript $m$ is omitted, further details on multiple regression tasks are provided in the supplements \footnote{\scriptsize{\url{http://people.duke.edu/~yz196/pdf/AAAI_suppl.zip}}}.

\begin{figure}[t!]
	\centering
	\resizebox{200px}{!}{\includegraphics{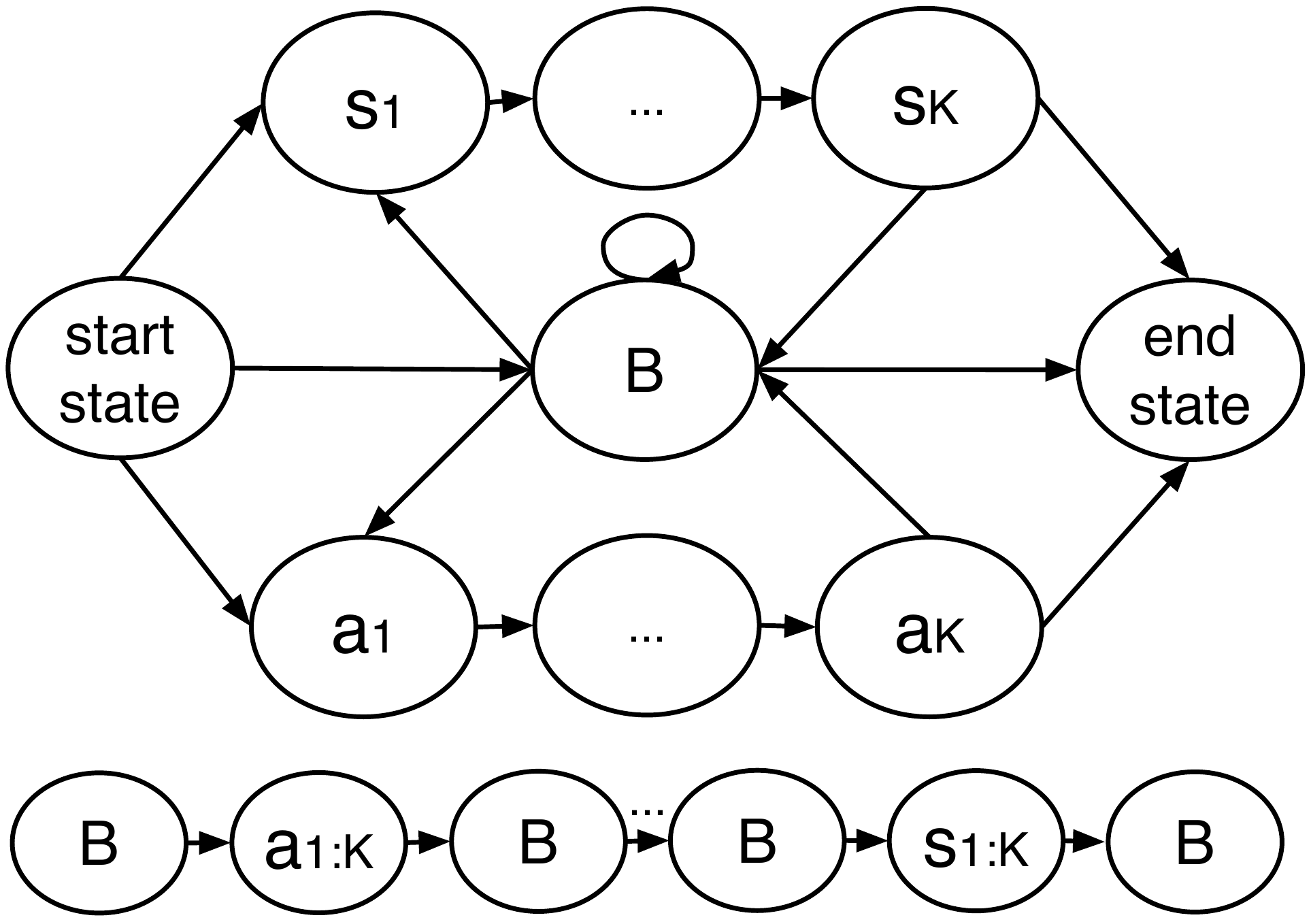}}
	\caption{(Top) PBM setup: background state can either transit to itself, or to a series of motif states $a_1 \rightarrow a_K$ or $s_1 \rightarrow s_K$. (Bottom) Example of a hidden states path.}\label{f2}
\end{figure}

In this scenario, we have the background (non-binding) hidden states denoted as $B$, and the motif hidden states $a_{1},\ldots,a_{K}$ and $s_{1},\ldots,s_{K}$ indicating being bound by an assayed transcription factor, where $K$ denotes the motif length. Motifs can be represented as sense or antisense, $s_{1} \rightarrow\ldots\rightarrow s_{K}$ or $a_{1} \rightarrow \ldots \rightarrow a_{K}$, respectively. Hidden variables are enforced to traverse through all $K$ sub-sequential motif states, once entering a motif start state $s_{1}$ or $a_{1}$, as shown in Figure~\ref{f2}. Corresponding states, $s_{k}$ and $a_{K-k}$, have complementary emissions, i.e., $p(x =A|z = s_{k}) = p(x =T|z = a_{K-k})$, $p(x =C|z = s_{k}) = p(x = G|z = a_{K-k})$. The transition probabilities from background states to motif start states, defined as $A_{B\rightarrow m} \triangleq A_{B\rightarrow a_1}+A_{B\rightarrow s_1}$, characterize the \emph{prior} information on the frequency of binding. In a real scenario, this frequency may vary along the probe because the free end of the probe is more susceptible to being bound \cite{Orenstein:2014dm}. To model this \emph{positional bias}, we introduce a position-dependent transition, specifically,
\begin{align*}
 	 P ( z_{l+1}=\{s_1,a_1,B\} | z_{l} =B) = \bm{A}^{(l)}_{B\rightarrow \{s_1,a_1,B\} } \,,
\end{align*}
where $l=1,\ldots,L$ indicates the position along the sequence. Note that in the PBM setting, the elements of transition matrix $A$ other than $A_{B\rightarrow s_1}$, $A_{B\rightarrow a_1}$ and $A_{B\rightarrow B}$ are deterministic (either $0$ or $1$). Thus, only $A^{(l)}_{B\rightarrow s_1}$, $A^{(l)}_{B\rightarrow a_1}$ and $A^{(l)}_{B\rightarrow B}$ can vary along the  sequences. 

To align with the fact that signal intensities reflect the number of TFs bound, we explicitly choose the summary function, $g(\cdot)$, to count the total occurrences of a certain set of states, $\mathbb{S}$, on each sequence, i.e.,
\begin{eqnarray*}
  v_i = g(\bm{Z_i}) = \sum_{l=1}^{L} I ( z_{il} \in \mathbb{S} ) \,,
\end{eqnarray*}
where $I(\cdot)$ is the indicator function and $\mathbb{S}\in{1\ldots S}$ is the collection of states of interest. Three possible settings for $\mathbb{S}$ are examined: counting the motif ``entering'' states, $\mathbb{S}=\{s_{1},a_{1}\}$, counting the motif ``exiting'' states, $\mathbb{S}=\{s_{K},a_{K}\}$ and counting the ``total bound'' states, $\mathbb{S}=\{s_{1,\ldots,K},a_{1,\ldots,K}\}$.

The link function itself, $f(\cdot)$, can be either linear or hyperbolic tangent. The motivation for a non-linear link comes from the observation that probes can ``saturate'' with increasing binding instances, leading to a nonlinear association between number of bound TFs and signal intensity responses \cite{Berger:2009jq}. Empirically, the nonlinear hyperbolic tangent link outperforms a linear one. When a non-linear transformation via tanh is used, the mapping from hidden space to responses can be understood as sigmoid belief network (SBN) with a single continuous output and constrained weights.
Comparisons between different summary and link functions are provided in the supplements.

\subsection{EM inference for coupled regression HMM model}
Suppose we work on PBM data, where our goal is to use expectation maximization (EM) to maximize joint likelihood $p ( \bm{X},\bm{Y} |\bm{\Theta},\bm{\Gamma})$ with respect to HMM parameters $\bm{\Theta}$ and regression parameters $\bm{\Gamma}$. A review and comparison with the standard Baum-Welch algorithm, which considers maximizing $p(\bm{X}|\bm{\Theta})$ can be found in the supplements.

In {\emph{E-step}}, we estimate the conditional distribution of hidden state sequence $\bm{Z}_{i}$ for $i$-th sequence, then we calculate the $Q$ function, i.e., the expected value of the log complete likelihood as
\begin{align}
  & Q(\bm{\Theta},\bm{\Gamma})  = \mathbb{E}_{\bm{Z} |
  \bm{X},\bm{Y}, \bm{\Theta}^{s} , \bm{\Gamma}^{s} }   [\log  p (
  \bm{X},\bm{Y},\bm{Z};\bm{\Theta},\bm{\Gamma})] \label{eq:Qe} \\
  & = \mathbb{E}_{\bm{Z} | \bm{X},\bm{Y},\bm{\Theta}^{s},
  \bm{\Gamma}^{s}} [ \log p (\bm{X},\bm{Z};\bm{\Theta})  
   + \log p (\bm{Y}|\bm{Z};\bm{\Gamma})] \nonumber \\
  & = Q_{1} (\bm{\Theta}) + Q_{2} (\bm{\Gamma}) \,, \label{eq:Qm}
\end{align}
\noindent where $\{\bm{\Theta}^{s},\bm{\Gamma}^{s}\}$ denote the model parameters estimated by previous step, $s$. The expectation in~\eqref{eq:Qe} can be factorized provided that $\bm{X}$ and $\bm{Y}$ are conditionally independent given $\bm{Z}$. In \emph{M-step}, $Q_{1}(\bm{\Theta})$ and $Q_{2}(\bm{\Gamma})$ from~\eqref{eq:Qm} are then maximized.

We first examine $Q_{1}(\bm{\Theta})$, a key observation is that this expectation can be reformulated as the expectation with respect to $P( z_{il} | \bm{X}_{i} ,\bm{Y}_{i} ,\bm{\Theta}^{s},\bm{\Gamma}^{s})$ or $P ( z_{il} ,z_{i ( l+1 )  } | \bm{X}_{i} ,\bm{Y}_{i},\bm{\Theta}^{s},\bm{\Gamma}^{s} )$. Thereby, HMM parameters $\bm{\Theta}$ can be updated with procedures similar to Baum-Welch. The only difference lies in the fact that in standard Baum-Welch, $P(z_{il}|\bm{X}_{i},\bm{\Theta}^{s})$ can be fast and accurately computed via the forward-backward algorithm, whereas in our case, the conditional distribution of hidden variables depends on both $\bm{X}$ and $\bm{Y}$, preventing us from directly applying forward-backward filtering. As we discussed in the section above, the transition probabilities can vary along the sequence. This can be easily
handled via a position-wise update of transition parameters. 

Denoting $\bm{X}_{i}$ to be the $i$-th sequence. For $i$-th input,
\begin{align*}
  & p ( z_{il} | y_{i} , \Gamma^{s} ,\bm{X}_{i} , \bm{\Theta}^{s}
   ) \\
  & \hspace{8mm} \propto \ p ( z_{il} | \bm{X}_{i} , \bm{\Theta}^{s}
   ) p ( y_{i} | z_{il} ,\bm{X}_{i} , \Gamma^{s} , \bf{\Theta}^{s}
   ) \\
  & \hspace{8mm} \propto \ p ( z_{il} | \bm{X}_{i} , \bm{\Theta}^{s}  )
  \sum_{v_{i}} p ( y_{i} | v_{i} , \Gamma^{s}  ) p ( v_{i} |
   z_{il} ,\bm{X}_{i} , \bf{\Theta}^{s} ) \,,
\end{align*}
where $p ( z_{il} | \bm{X}_{i},\bm{\Theta}^{s})$ and $p(y_{i} | v_{i} ,\bm{\Gamma}^{s})$ can be cheaply obtained. However, the conditional distributions over hidden summaries, i.e., $p (v_{i} | z_{il} ,\bm{X}_{i} ,\bm{\Theta}^{s})$, are not straightforward. A naive approach would utilize a Gaussian distribution to approximate the true posterior, where mean and variances of the Gaussian distribution can be estimated analytically using propagation tricks similar to forward-backward calculations, as detailed in the supplements. Unfortunately, this approach only works when the real valued distribution $p ( v_{i} | z_{il},\bm{X}_{i},\bm{\Theta}^{s})$ is unimodal, which is generally not the case.

Instead, we appeal to a Viterbi path integration approximation. The approximated distribution is achieved by integrating over a truncated hidden state sequence space, denoted here as $\bm{\mathbb{P}}_{\text{vit}}$, where the number of included Viterbi paths is dynamically determined to cover 99\% of the total probability mass, thus
\begin{align*}
  p ( v_{i} | z_{il} ,\bm{X}_{i} , \bm{\Theta}^{s})
  \propto \mathbb{E}_{\bm{Z}_{i} \in
  \bm{\mathbb{P}}_{\text{vit}}} p ( v_{i} | \bm{Z}_{i}
   ) p ( \bm{Z}_{i} | z_{il} ,\bm{X}_{i} , \bm{\Theta}^{s}
   ) \,,
\end{align*}
where the $\bm{Z}_{i}$ are the Viterbi paths conditioned on $\bm{X}_{i}$ and $z_{il}$, and $p(v_{i} | \bm{Z}_{i})$ is an indicator function $I(v_{i} =g ( \bm{Z}_{i} ) )$, provided that summaries $v_i$ are deterministic given a hidden state sequence $Z_i$. The calculation of $p(\bm{Z}_{i} | z_{il} =z' ,\bm{X}_{i},\bm{\Theta}^{s})$ is performed for each position $l$ and each possible state $z'$ that $z_{il}$ can take. For each $\{l,z'\}$ pair, the computational complexity is ${\cal O}(S^2 L)$, thus a naive approach would cost ${\cal O}(S^3 L^2)$. To reduce computational cost we adopted a dynamic programming approach described in Algorithm~\ref{alg1}, in which no additional cost is required. Briefly, this procedure first propagates through the sequence to calculate a forward and a backward tensor, while keeping track of the top paths, then it computes the conditional probabilities for each $\{l,z'\}$ pair altogether. In Algorithm~\ref{alg1}, the symbol $\otimes$ represents the Kronecker product and we use $D$ to denote the number of top paths.
\begin{algorithm}[t!]
\caption{Top ${D}$ Conditional Viterbi Path Integration.}\label{alg1}
\begin{algorithmic}[1]
\STATE \emph{(For each $(l,z')$ combination, estimate the distribution of summary variable $v_i$ conditioned on $z_{il}=z'$)}	
\STATE Initialize forward tensor at position 0, $F( 0,1 \colon {S},1 \colon {D} )$    
\FOR {$l=1  \text{to}  L $}
\FOR {$z'=1  \text{to}  S $}
\STATE $F ( l,z', 1 \colon {D} ) = \text{MaxK} (F ( l-1,1 \colon {S}  ,1 \colon {D} )    
+A ( 1 \colon {S} ,z' ) ) +E ( z',x_{il} )$
\ENDFOR
\STATE{\emph{(where MaxK($\cdot$) is a routine to find top ${D}$ elements in a matrix via partial sorting)}}
\ENDFOR
\STATE Initialize backward tensor at pos $L$, $B( L,1 \colon {S},1 \colon {D}) $
\FOR {$l=L-1  \text{to}  1 $}
\FOR {$z'=1  \text{to}  S $}
\STATE $B ( l,z',1 \colon {D} ) = \text{MaxK} ( B ( l+1,1 \colon {S} ,1 \colon {D} ) +A
( z' ,1 \colon {S} ) \} +E ( z',X_{l} )$
\ENDFOR
\ENDFOR
\\
\FOR {$l=1  \text{to}  L $}
\FOR {$z'=1  \text{to}  S $}
\STATE $\text{C} ( l,z',1 \colon {D} ) = \text{MaxK}(F ( l,z',1 \colon {D} ) \otimes B ( l,z',1 \colon {D} ) -E ( z',x_{il} ) )$
\STATE{\emph{(where $C(l,z',k)$ represents the probability of top $k$th conditional Viterbi path)}}
\STATE Calculate summary variable $V(l,z',1 \colon {D})$
\ENDFOR
\ENDFOR
\IF {Any $\sum_{k=1}^{D}{C(l,z',k)} < \text{threshold}$}
\STATE Dynamically increase $D$, redo the forward-backward procedure
\ELSE 
\STATE Renormalize $C$ 
\ENDIF
\STATE Return C,V
\end{algorithmic}
\end{algorithm}

In the worst case, Viterbi path integration may require a large number of top Viterbi paths, which could be computationally prohibitive. However, in our experiments on PBM data, around 100 Viterbi paths usually cover 99\% probability mass for most probe sequences.

So far we have considered maximizing the $Q_{1}(\bm{\Theta})$ with Viterbi path integration. We can rewrite $Q_{2} (\bm{\Gamma})$ as
\begin{align}\label{eq:Q2}
  Q_{2}(\bm{\Gamma}) = \mathbb{E}_{\bm{V} |
  \bm{X},\bm{Y}, \Theta^{s} , \Gamma^{s} } \log  p (
  \bm{Y} |  \bm{V}, \bm{\Gamma},\sigma ) \,.
\end{align}
Maximizing $Q_{2} (\bm{\Gamma})$ can be considered as a least square regression task with random variable covariates, $\bm{V}$. If the link function between $\bm{Y}$ and $\bm{V}$ is chosen to be linear, an analytic solution based on \emph{expected summary sufficient statistics} (ESSS), $\mathbb{E}_{p ( v_{i} | \bm{X}_{i},y_{i} , \bm{\Theta}^{s} , \bm{\Gamma}^{s}  )} v_{i}$ and $\mathbb{E}_{p (v_{i} | \bm{X}_{i} ,y_{i} , \bm{\Theta}^{s} , \bm{\Gamma}^{s}  )} v_{i}^{2}$ can be computed. For a tanh link function, coordinate gradient descent (CGD) can be utilized to update $\alpha$ and $\beta$ analytically using ESSS. The shape and position parameters, $\{s,t\}$, can be updated using standard trust-region algorithm \cite{sorensen1982newton}. The expectation in~\eqref{eq:Q2} is with respect to $p(v_{i}|\bm{X}_{i},y_{i}, \bm{\Theta}^{s} , \bm{\Gamma}^{s}  )\propto p ( y_{i} | v_{i} , \bm{\Gamma}^{s}  ) p ( v_{i} | \bm{X}_{i},\bm{\Theta}^{s}  )$, which can be calculated using same Viterbi path integration approach described before.

Full details of EM inference are provided in the supplements. MCMC approaches such as Gibbs sampling with embedded Metropolis-Hastings can be applied to this model. However, directly sampling hidden states may be negatively affected due to local modes, rendering mixing arbitrarily slow. Inherent forward-backward calculation in EM may alleviate this problem. Backpropagation may not be appropriate for two reasons. First, in our model the hidden variables generate both $\bm{X}$ and $\bm{Y}$, where backpropagation approaches are not straightforward. Second, we wish to estimate a distribution over hidden state sequences rather than a point estimate. Variational inference methods such as mean field or variational auto-encoders \cite{Kingma:2013tz,Fabius:2014uo} might be possible, but at the cost of accuracy and implementation complexity. Benefitting from caching and dynamic programing, we achieve time complexity of ${\cal O}(NS^{2}L)$ multiplied by the number of iterations, which is essentially the same as in Baum-Welch.

\section{Experiments}
\subsection{Simulation studies}
We conducted a simulation study in a PBM scenario, where we are interested in recovering the correct emission matrix and predicting for new observations. As a baseline, we considered a two-stage approach, where $\bm{\Theta}$ is first estimated with Baum-Welch algorithm, followed by a regression task to estimate $\bm{\Gamma}$. In our synthetic PBM data, 20,000 sequences (sequence length $L=20$) and corresponding signals were generated from the generative model parameterized by $\bm{\Theta}$ and $\bm{\Gamma}$. 10,000 sequence instances were used for training, while the remaining 10,000 instances were used for evaluation. A motif of length 5 with both sense and antisense representation was used. To make the task more challenging, the transition probability from background state to motif states was set to be relatively small, so that motif states were not enriched. We examined 6 different setups for combinations of summary functions (all states, entering states, exiting states) and link functions (linear, tanh). For clarity, we only show the results of one setting (all state + linear), as the other 5 settings gave similar results. Detailed results are provided in the supplements.

\begin{figure*}[t!]
	\centering
	\subfigure[]{\includegraphics[width=40mm,trim=0 0 0 0 mm,clip]{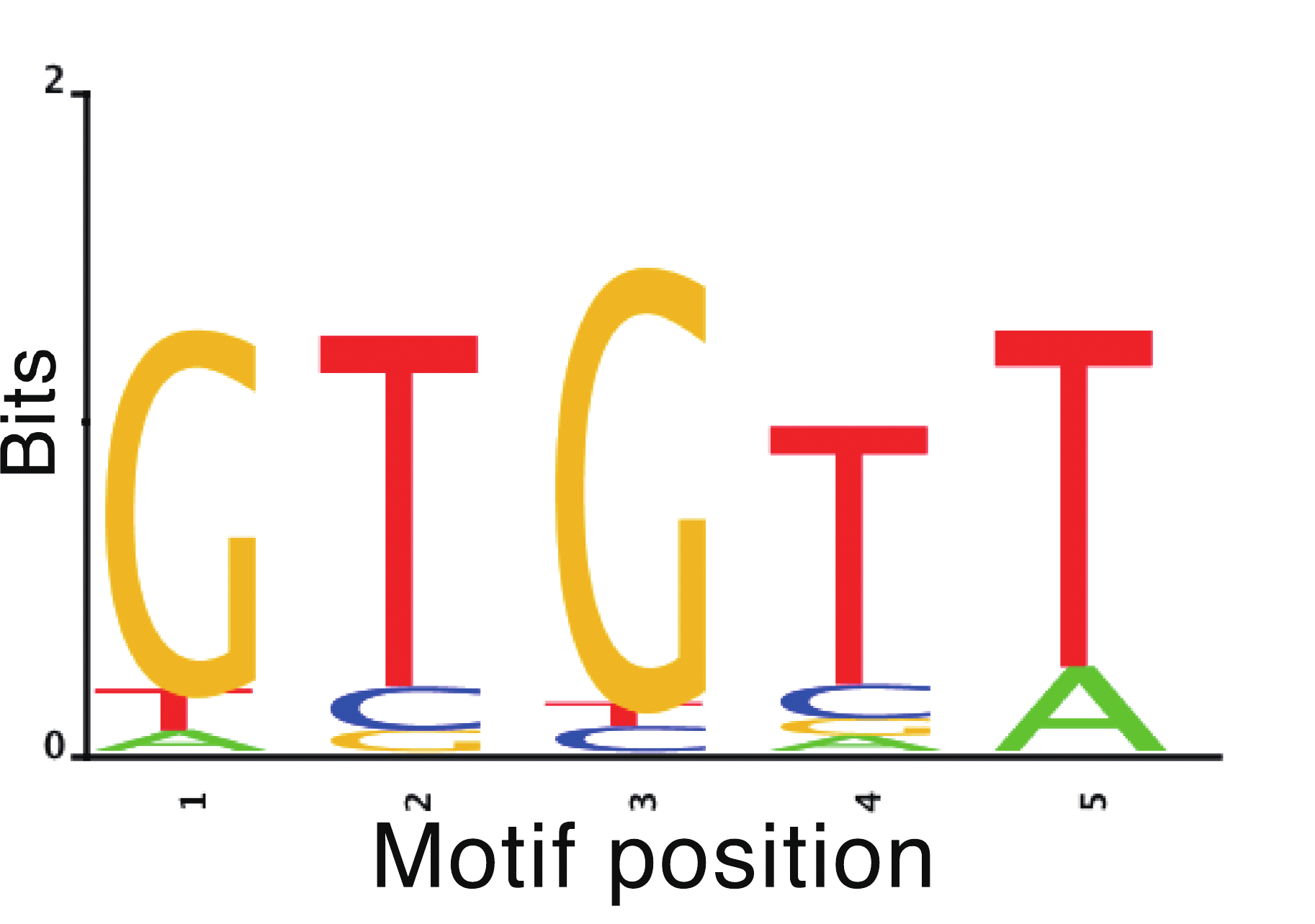}} 	\subfigure[]{\includegraphics[width=40mm,trim=0 0 0 0 mm,clip]{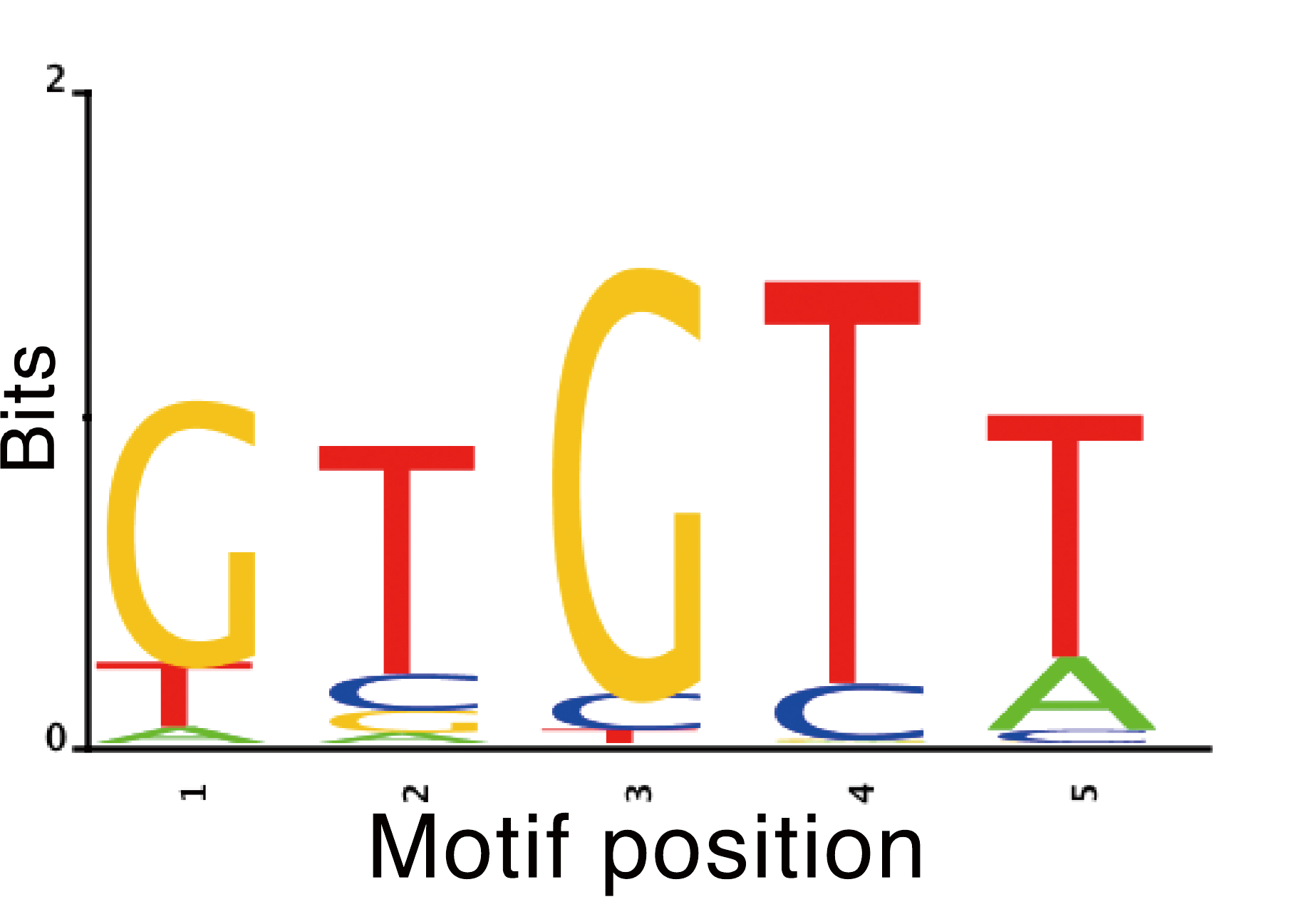}} 	\subfigure[]{\includegraphics[width=40mm,trim=0 0 0 0 mm,clip]{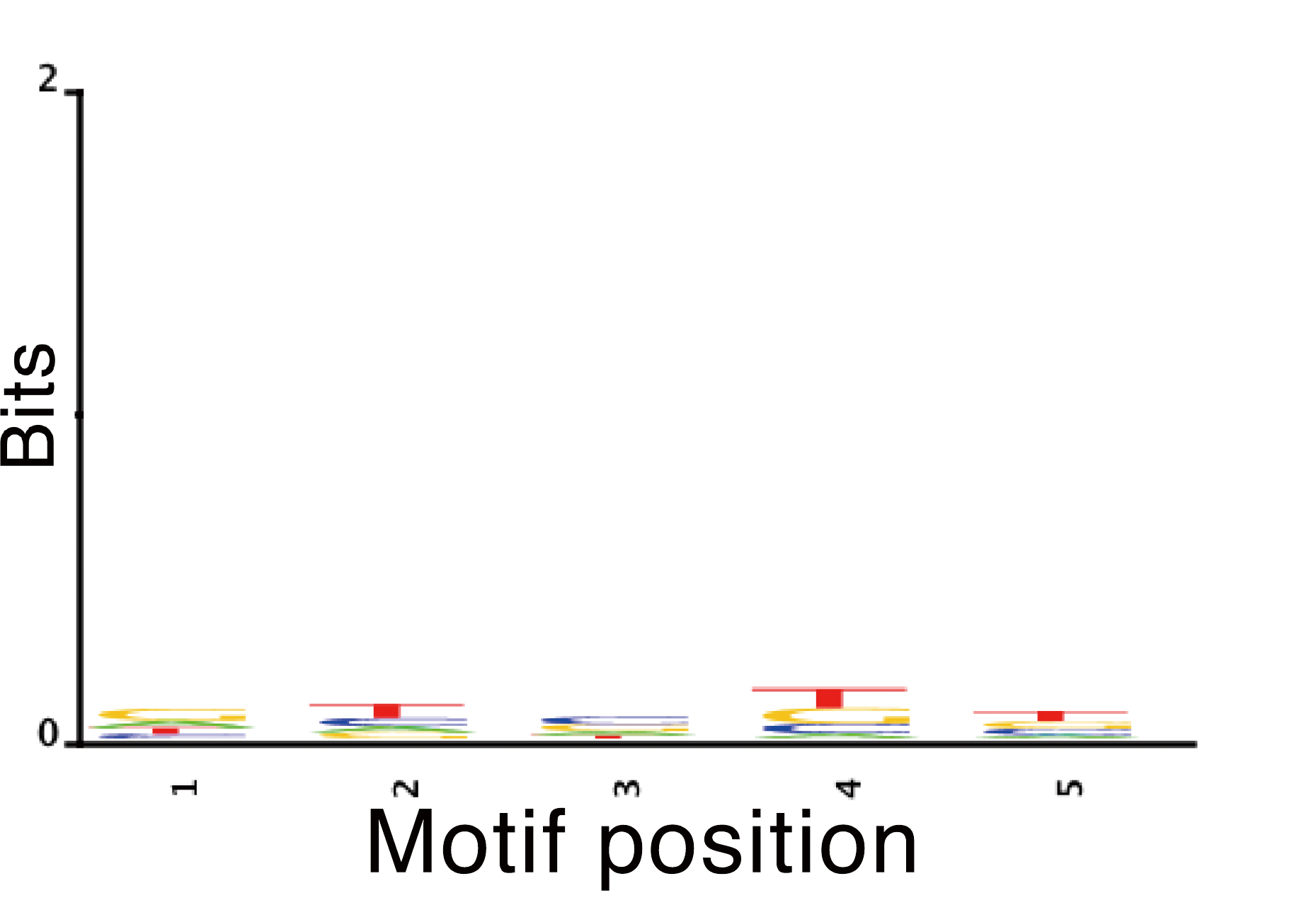}}
	\subfigure[]{\includegraphics[width=40mm]{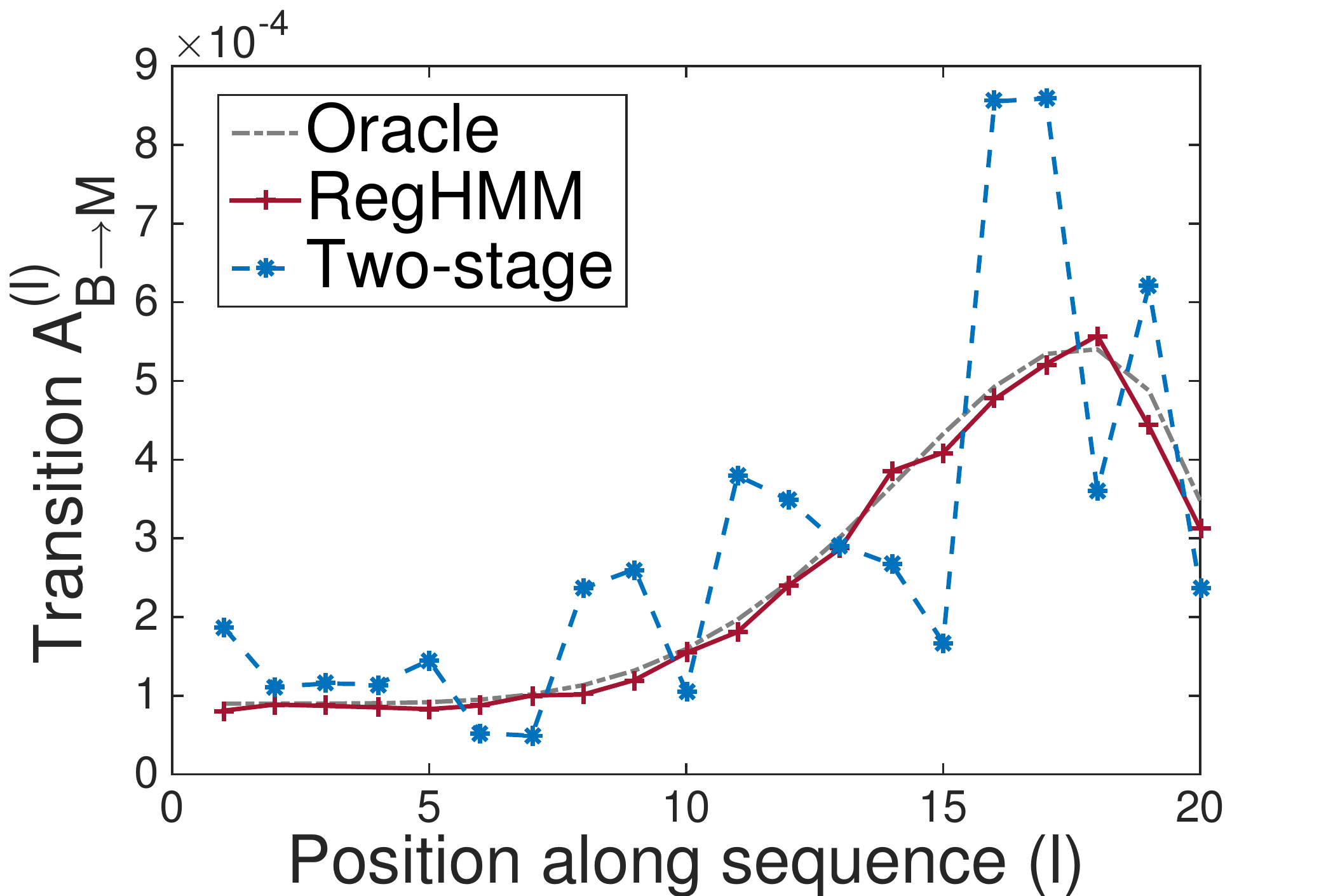}} 
	\caption{Inferred HMM parameters from simulated data. a): Ground truth. b-c): Motif patterns learned by RegHMM and two-stage model, respectively. d): Inferred transition probabilities along the probe, from background state to motif state $A_{B\rightarrow M}^{(l )}$, where $A_{B\rightarrow M}^{(l )} \triangleq p ( z_{l} = \text{motif} | z_{l-1} = \text{background} )$. }\label{f3}
\end{figure*}

Table~\ref{t1} shows that our method outperformed two-stage baseline model in three aspects. First, the coupled model could recover parameters $\bm{\Theta}$ and $\bm{\Gamma}$ close to the truth, whereas the two-stage model diverged significantly. The PWM model implied by emission matrix is shown in Figure~\ref{f3}, RegHMM could capture the emission and transition probability well even under the case where sequential observations were less informative. Note that we were only interested in background to motif transition $A_{B\rightarrow M}^{(l)}$. Second, mean squared prediction error for held-out responses (prediction MSE) from test sequences was lower. Finally, as a generative model, the likelihood of unseens (test log-likelihood) was higher. These results indicate that our model can bridge the information between sequential observations and real responses, and leverages all the information to parse the sequences correctly. Having a good understanding about data, the algorithm can then predict reasonably well. To see how the additional real responses reinforce the learning on HMM side, intuitively, these extra responses function via both re-weighting contribution of different sequences and highlighting individual positions in each sequence according to consistency between predicted responses and observed responses. Allowing transitions to vary along the sequence yield slight improvement of performances (see supplements).
\begin{table}[t!]
  \begin{center}
	\setlength{\tabcolsep}{2pt}
	\begin{tabular}{cccc}
      \hline
      & $\text{Oracle}$ & $\text{RegHMM}$ & $\text{Two-stage}$\\
      \hline
      $\alpha$ & 5 & 5.00 & 4.90\\
      $\beta$ & 4 & 4.10 & 20.61\\
      $\| \hat{\bm{E}} -\bm{E} \|_{\infty}$ & 0 & 0.19 & 1.16\\
	  $\| \hat{\bm{A}}_{B \rightarrow M} -\bm{A}_{B \rightarrow M} \|_{\infty}$ & 0 & $4.5\times10^{-5}$ & $3.6\times10^{-4}$\\
	  $\text{Pred. MSE}$ & 0.962 & 0.969 & 1.355 \\
      $\text{Test LL}$ & $-29.13$ & $-29.14$ & $-29.81$
	\end{tabular}
  \end{center}
  \caption{\label{t1}Comparison between RegHMM and baseline (two-stage) model on synthetic data.}
\end{table}

We also simulated a case where the observed sequences cannot be distinguished from sequences generated by a null model, i.e., where only background state transmits to itself. Remarkably, RegHMM recovered the correct motif pattern even when HMM structure was seemingly missing, whereas two-stage model failed (see supplements).

\subsection{Real data analysis}
We evaluated our algorithm on DREAM5, a publicly available dataset of PBM experiments for methods competition \cite{Weirauch:2013ju}. The DREAM5 dataset consists of experimental data for 66 transcription factors under two array designs. The overall task is to use one array type ($\sim41,000$ probe sequences) for training and predict the real signal intensity responses on the other type. Apart from the prediction task, the inferred motif patterns are also of interest.

We compared RegHMM with BEEML \cite{Zhao:2011isa}, Seed and Wobble \cite{Berger:2006fv}, RankMOTIF++ \cite{Chen:2007ed}, FeatureREDUCE \cite{Weirauch:2013ju} and DeepBind \cite{Alipanahi:2015fba}. As described in Weirauch et al., all sequences were tailored to 35 bases to retain only unique sequences ($L=35$). Also, original signal intensities were first transformed using a logarithmic function \cite{Weirauch:2013ju}. For RegHMM, we considered both linear and tanh link functions and two summary functions: counting all motif states and count motif entering states. We also compared between models with and without position-wise transitions. The motif length $K$ for all scenarios was set to 6. The initial value for the emission matrix was established from the most frequent 6-mer.

Following Weirauch et al., the comparison was conducted based on two metrics. The Pearson correlation coefficient between predicted responses and observed responses, to assess the predictive ability of each model for unseen queries. Apart from this, we also compared the area under the receiver operating characteristic (AUROC), as the experimental data can be subject to noise and outliers. To align with the comparison done by Weirauch et al., the positives were defined as the probes that are 4 standard deviations above the mean, with an enforced minimum of 50.

\begin{figure}[ht!]  
	\centering
	\resizebox{110px}{!}{\includegraphics{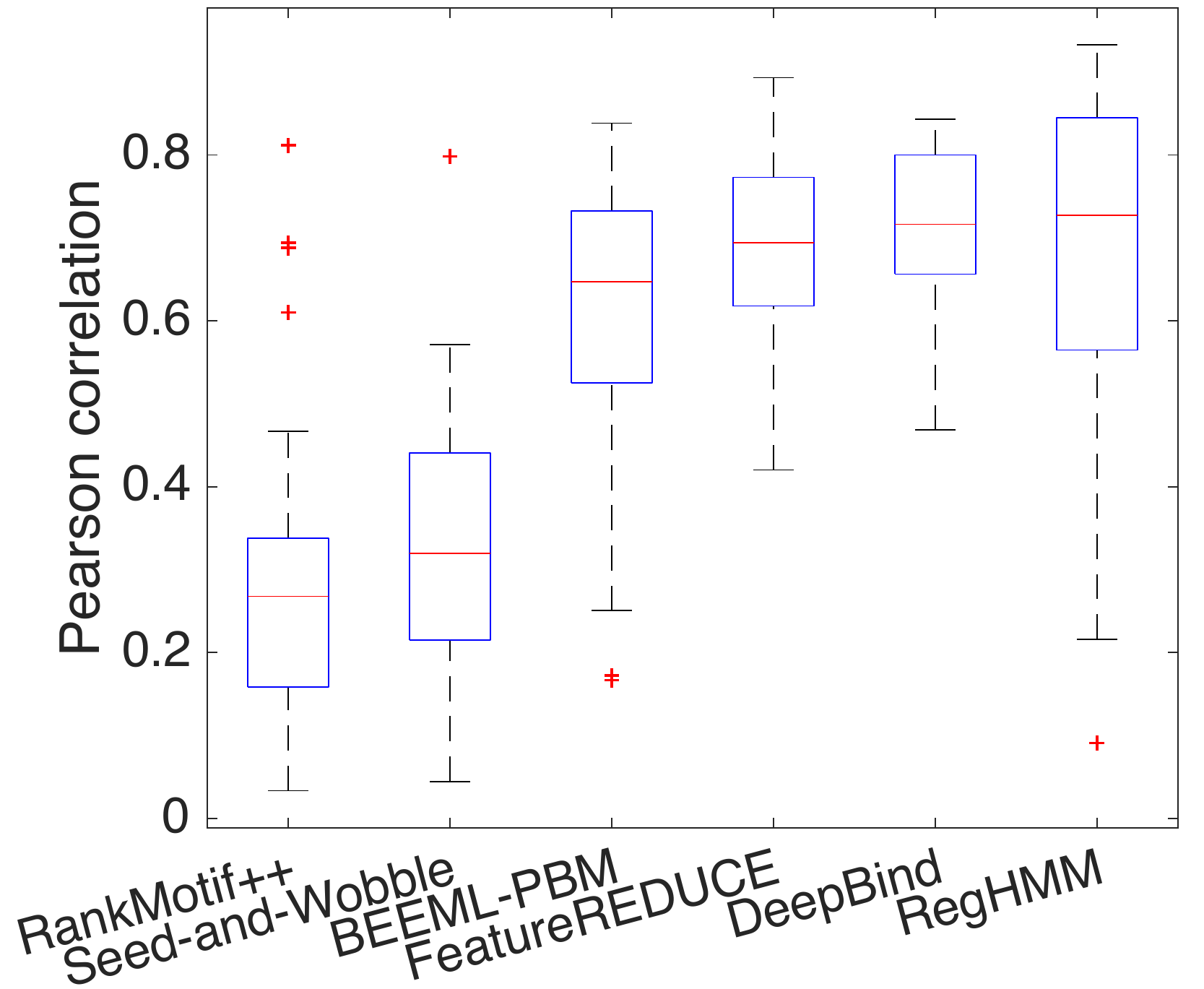}}\resizebox{110px}{!}{\includegraphics{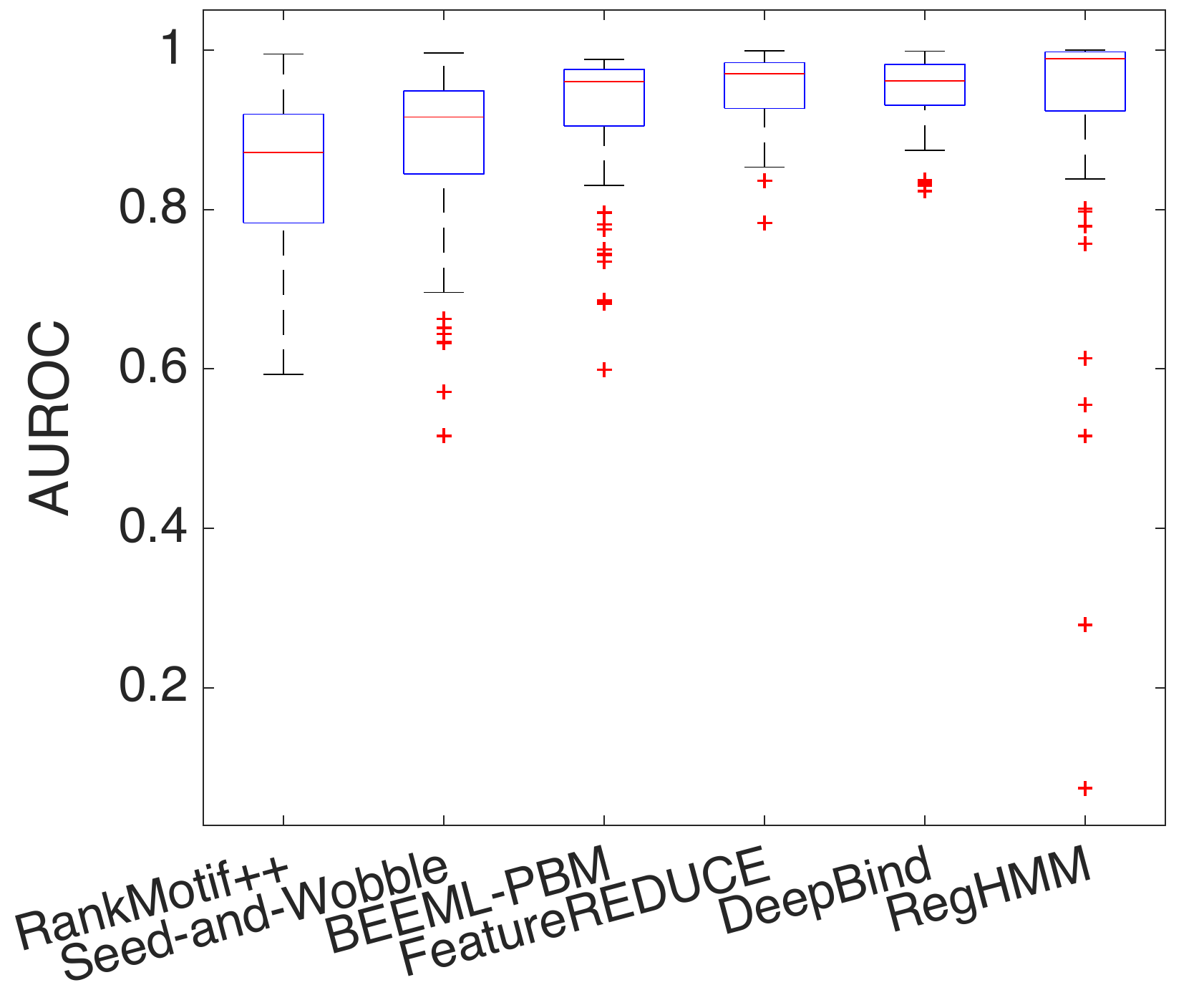}}
	\caption{DREAM5 probe level comparison based on correlation (left panel) and AUROC (right panel).}\label{t2}
\end{figure}
\begin{figure}[t!]
	\centering
    {\resizebox{115px}{!}{\includegraphics{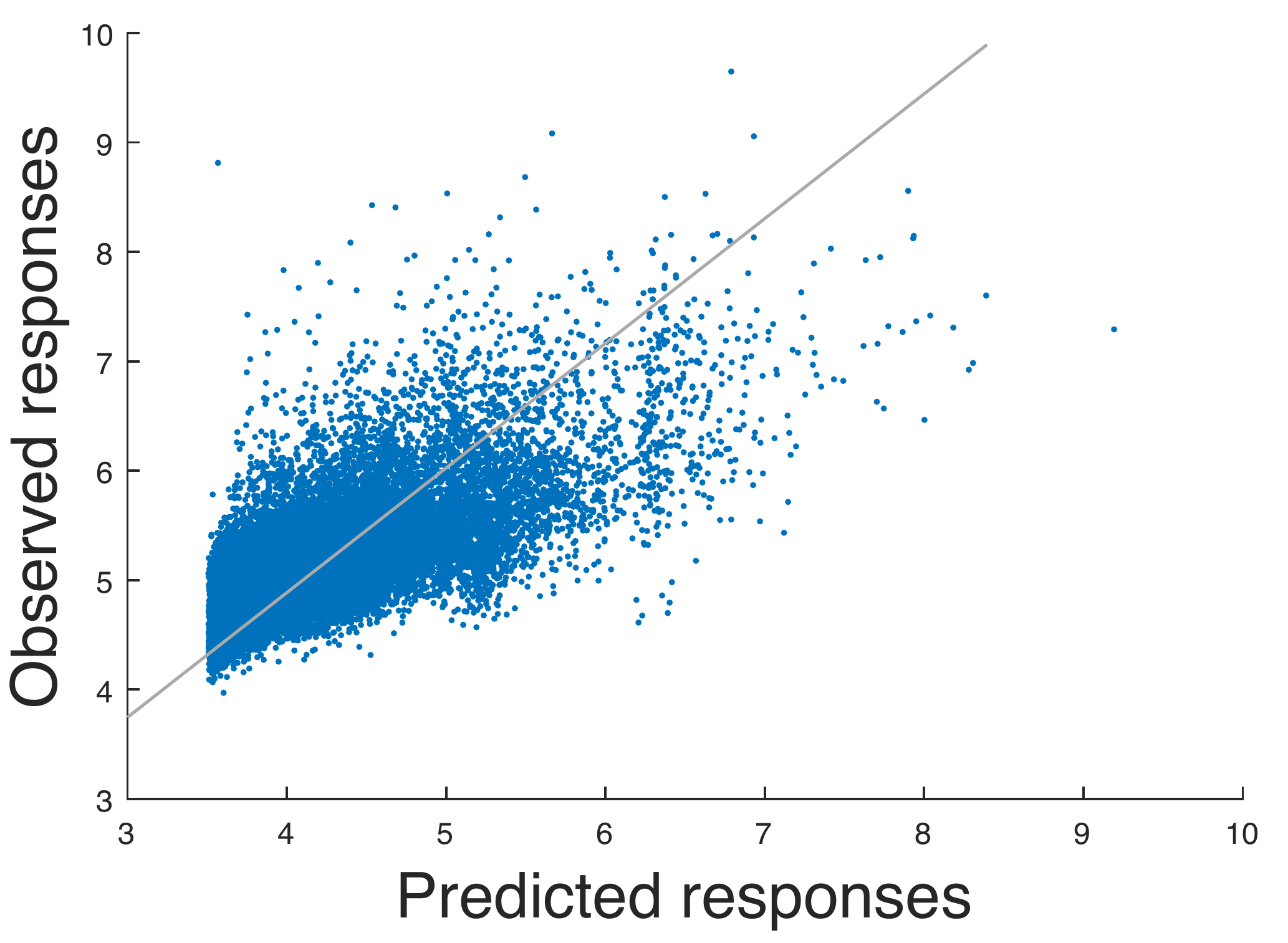}}} 
  \caption{Scatter plot of predicted responses and observed response of Zscan10. Pearson correlation $\rho = 0.82$.}\label{f5}
\end{figure}

As shown in Figure~\ref{t2}, RegHMM achieved comparable average Pearson correlation and AUROC compared with other methods over 66 experiments. We emphasize that our model can directly infer a motif pattern, whereas K-mer based models parse sequences into features that do not explicitly explain the data. Thus, our model is predicting ``reasonably'' by attempting to simultaneously interpret the data. Interestingly, for several transcription factors (such as TF-60, TF-66) reported to be hard to predict by other methods, our model achieved good prediction (figure~\ref{f5}). Unexpectedly, the link functions learned from some transcription factors are monotonically decreasing, indicating the number of motif instances is negatively correlated with signal intensity responses, which  may explain why other models assuming only positive associations fail. Possible explanations would be that the motif found captures sequence bias of PBM experiments to some extent, or the assayed transcription factor inclined to avoid binding to certain motif pattern.

Another observation was that if we train two models separately for each one of the two distinct array designs, even though the inferred regression parameters showed slightly discrepancy, the learned HMM parameters corresponded reasonably well, suggesting a good consistency between arrays. Moreover, we found the transition probability from background to motif states (which we defined as positional bias in previous sections) decreased with positional index (see supplements), which is consistent with our knowledge that free ends of the probe favor binding.

We compared different link functions and summary functions. As shown in the supplements, using a tanh link function marginally outperformed a linear one. The choice of summary function was more TF-specific. Nevertheless, for most TFs, a summary function considering all motif states led to better performance.

\section{Discussion}

Motivated by a TF motif discovery challenge, we examined the problem of learning a regression-coupled HMM. We proposed RegHMM, a probabilistic generative model learning hidden labeling of observed sequences by leveraging real-valued responses via a flexible set of link and summary functions. An approximate EM algorithm using Viterbi path integration is employed for fast and accurate inference. The experiments on simulated data showed that, by incorporating additional responses, RegHMM can accurately label the sequences and predict on future observations in situations where inference from sequential observation alone fails. The application to a real-world dataset showed that RegHMM can explicitly characterize protein-binding preferences, while maintaining predictive power. The results suggested potential biological insights for further validation. 

Our algorithm was designed for complementing sequential data with real-valued summaries. However, using a softmax or Poisson link, our model could be naturally extended to handle categorical or count responses. Applications may include motif finding from \emph{in vivo} data where multiple tracks of count responses are available. In addition, task-specific summary functions can be engineered to reflect domain knowledge. For example, indicator functions of motif orderings may associate with certain signal responses.

\clearpage

\section{Acknowledgments}

We thank Ning Shen from Duke university for helping process the data. This research was supported in part by NSF, ARO, DARPA, DOE, NGA and ONR.

\bibliographystyle{aaai}
\bibliography{ref}

\end{document}